\documentclass[final]{midl} % Include author names
\jmlryear{2019}
\jmlrproceedings{}{Pages}

\title{Adversarial Augmentation for Enhancing Classification of Mammography Images}

\midlauthor{\Name{Lukas Jendele\midljointauthortext{Contributed equally.}\nametag{$^{1}$}} \Email{jendelel@ethz.ch}\\
\and
\Name{Ondrej Skopek\midlotherjointauthor\nametag{$^{1}$}} \Email{oskopek@ethz.ch}\\
\addr $^{1}$ Department of Computer Science, ETH Z\"{u}rich \\
\AND
\Name{Anton S. Becker\nametag{$^{2,3}$}}
\Email{anton.becker@usz.ch}\\
\addr $^{2}$ Institute of Diagnostic and Interventional Radiology, University Hospital Z\"{u}rich\\
\addr $^{3}$ Department of Health Sciences and Technology, ETH Z\"{u}rich\\
\AND
\Name{Ender Konukoglu\nametag{$^{4}$}} \Email{ender.konukoglu@vision.ee.ethz.ch}\\
\addr $^{4}$ Computer Vision Laboratory,
ETH Z\"{u}rich
}

\usepackage[utf8]{inputenc} % allow utf-8 input
\usepackage[T1]{fontenc}    % use 8-bit T1 fonts
\usepackage{hyperref}       % hyperlinks
\usepackage{url}            % simple URL typesetting
\usepackage{booktabs}       % professional-quality tables
\usepackage{amsfonts}       % blackboard math symbols
\usepackage{amsmath}       % blackboard math symbols
\usepackage{nicefrac}       % compact symbols for 1/2, etc.
\usepackage{microtype}      % microtypography
\usepackage{graphicx}
\usepackage{multirow}
\usepackage{multicol}

\usepackage{color}

\newcommand{\calL}[1]{\mathcal{L}_{#1}}
\newcommand{\E}[2]{\mathrm{E}_{#1}\left[#2\right]}

\begin{document}
\maketitle

\begin{abstract}%
Supervised deep learning relies on the assumption that enough training data is available, which presents a problem for its application to several fields, like medical imaging. On the example of a binary image classification task (breast cancer recognition), we show that pretraining a generative model for meaningful image augmentation helps enhance the performance of the resulting classifier. By augmenting the data, performance on downstream classification tasks could be improved even with a relatively small training set. We show that this ``adversarial augmentation'' yields promising results compared to classical image augmentation on the example of breast cancer classification.
\end{abstract}

%%%%%%
\section{Introduction}
Deep learning in computer vision has achieved great results in the past few years \cite{imagenet, nvidia}. Most of these have been enabled by more computational power and large amounts of data. Unfortunately, in many scientific fields such as medical imaging, there are usually several orders of magnitude fewer data samples to work with than in large-scale computer vision datasets. Leaving aside issues like anonymization and privacy, this poses several specific problems for anyone wishing to use medical imaging datasets:
\begin{enumerate}
    \itemsep0em
    \item Scarcity --- Data is hard to obtain, usually only a few samples are available per dataset.
    
    \item Bias --- Medical and other small datasets usually contain many more negative (healthy) images than positive ones (with a valid and confirmed illness). The reason for that is that the data usually comes from a real-world diagnostic process, where data is obtained even at a low suspicion threshold, since the potential harms of the imaging procedure are far outweighed by the benefit of a prompt diagnosis. Furthermore, in screening settings a large population of completely symptom-free subjects is deliberately examined.
    
    Often present are also ``confirmation'' images -- for a patient with a positive finding, many more images will be made to confirm the diagnosis and monitor the progress. This only increases the bias, as the dataset then has several positive images of the same patient. In other fields, variations on these processes also exist, all resulting in a similar bias.
    
    \item Noise --- Introduced by capturing devices, errors made during data processing or storage, or from a naturally noisy population (e.g.~synthetic implants, marker wires, or prior surgery related to the illness).
\end{enumerate}
All three of these issues pose a significant challenge for training classification models. In this work, we aim to partially alleviate the first two problems in the context of binary image classification. Our contributions are the following:
\begin{enumerate}
    \itemsep0em
    \item We train a generative model that has the ability to transform data from one class to the other and back with a CycleGAN architecture \cite{cyclegan}.
    \item We show that the classifier is partially fooled into thinking that the transformed images are of the respective real class-label distributions.
    \item We show that the performance of the classifier may improve when its training data is augmented with the transformed images, in comparison to classical image augmentation.
\end{enumerate}

\section{Related work}

Generative Adversarial Networks (GANs), proposed by \citet{gan}, have shown great potential for generating or modifying images.
Many studies focused on image augmentation using GANs \cite{apple,ganhands,robots}. The application to the medical domain is logical, because it is generally difficult to obtain data there, and all datasets are naturally heavily imbalanced. \citet{nvidia_mri} focuses on brain MRI augmentation using paired image-to-image translation similar to the pix2pix approach \cite{pix2pix}.

However, paired images (e.g.~the same breast in the same view with and without cancer) are very hard to obtain. Thus, we focus on unpaired image augmentation. In their work on CycleGAN, \citet{cyclegan} used a pair of GANs coupled with a cycle-consistent loss for unpaired image-to-image translation, and succeed in converting images between two domains (e.g.~horses to zebras). In our work, we adopt this idea to generate cancerous features into or remove them from mammography images.

Parallel to our research, \citet{lesionremoval} have applied the CycleGAN architecture to augment brain and liver MRI scans. Aligned with our work, they show that such augmentation boosts the classifier's performance.

For the actual cancerous lesion detection, there have been several studies utilizing deep neural networks on image patches, like \citet{levy2016breast}.
The reason for looking at smaller patches is mostly because of dimensionality reduction. There have also been attempts at detection by training on whole images \cite{dezso, shen2017end, hussain2017differential}. They all augment the dataset by translating, rotating, or flipping the images to improve the system's performance, which we also compare to in our experiments.

\section{Model}

Our approach consists of two models, trained separately. In the first step, we train a specific GAN architecture to learn a transformation from the domain of images of one class label to the domain of images of the other class labels. In the second step, we use the generative model to augment a Faster R-CNN classifier \cite{frcnn} to improve its performance.

\subsection{Generative augmentation model}

\begin{figure}[tb]
    \centering
    \subfigure[Healthy $\to$ cancer]{
        \centering
        \includegraphics[width=0.25\textwidth]{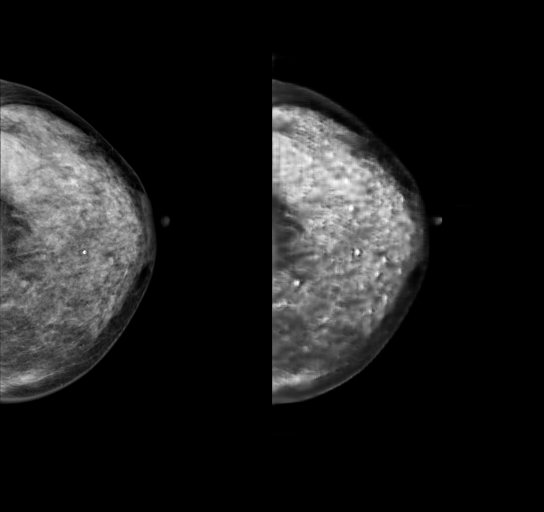}
    }
    \quad
    \subfigure[Cancer $\to$ healthy]{
        \centering
        \includegraphics[width=0.25\textwidth]{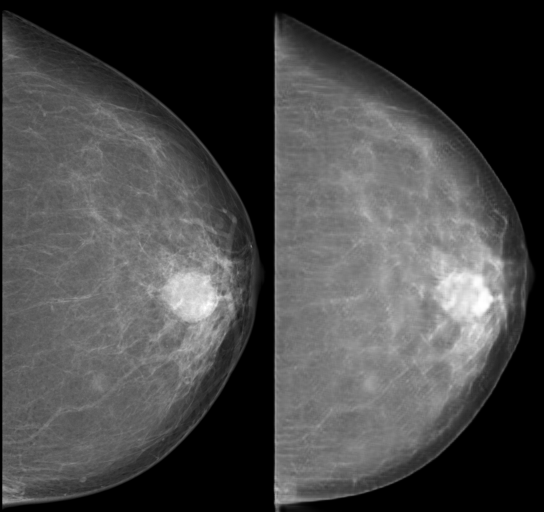}
    }
    \caption{Given an image dataset with two classes (cancerous and healthy breast scans), the generative model learns to transform images from one class to the other.}
    \label{fig:intro}
\end{figure}

The generative model is based on CycleGAN \cite{cyclegan}. Its goal is to perform unpaired translation of images from one domain to another, and back. It achieves this by training two generator--discriminator pairs and introducing a cycle-consistency loss. In our case, we apply it to generate and remove cancerous features from mammography images. 
Figure~\ref{fig:intro} shows the output of the generative model on two training samples.

More formally, CycleGAN transforms images from a domain $X$ to another domain $Y$. For that, it uses two independent mappings, $G_Y: X \to Y$ and $G_X: Y \to X$. To train these mappings directly, one would need paired images, which are very hard to obtain (for example, the same patient's image with and without breast cancer, in the exact same orientation).
Instead, CycleGAN uses a GAN-like loss of introducing a discriminator, that attempts to differentiate generated images from the empirical domain $\hat{X} = G_X(Y)$ from the real images from $X$ by learning a mapping $D_Y: (X \cup \hat{X}) \to [0, 1]$ (analogously for $D_X$).

\begin{figure}[tb]
    \centering
    \includegraphics[width=0.65\textwidth]{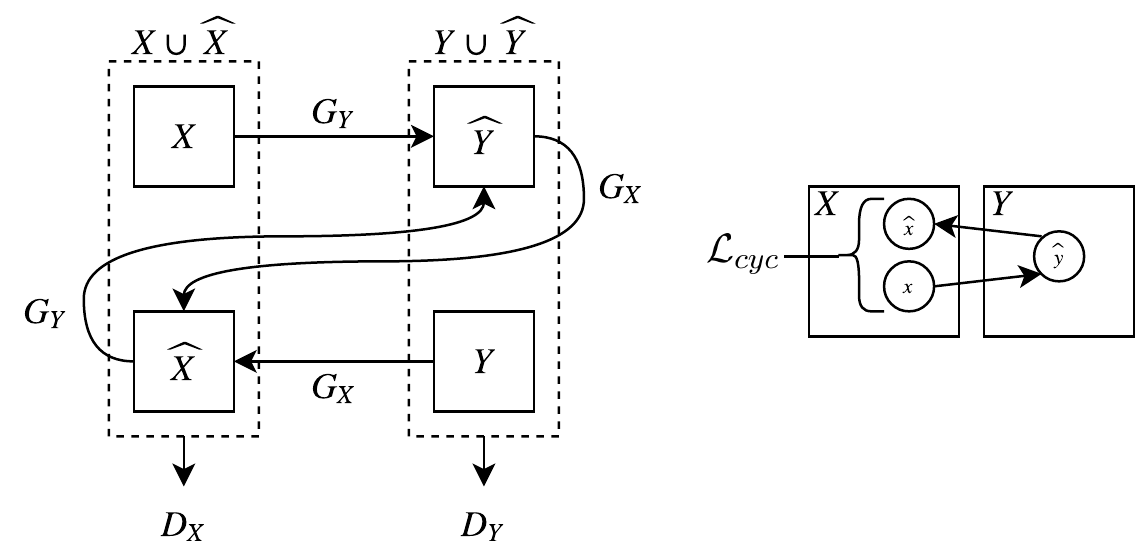}
    \caption{CycleGAN model diagram with the cycle-consistency loss $\calL{cyc}$, only shown $X\to Y$.}
    \label{fig:cyclegan}
\end{figure}

Furthermore, it adds a cycle-consistency loss $\calL{cyc}$, which enforces the "identity" property $X \approx G_X(G_Y(X))$. All of this is analogously done for domain $Y$ as well. Figure~\ref{fig:cyclegan} shows a simple diagram of the model.

The loss is composed of the following partial loss terms. The first is the classic adversarial GAN loss, where $D_X$ is the discriminator of the GAN on domain $X$, and $G_X$ is the generator of samples in the domain $X$ given a sample from $Y$.
$$\calL{GAN}(G_Y, D_Y) = \E{y \sim p_{data}(y)}{\log{D_Y(x)}} + \E{x \sim p_{data}(x)}{\log{\left(1-D_Y(G_Y(x))\right)}}$$
For training stability reasons, our implementation uses the alternative LSGAN \cite{lsq_gan} loss function, with parameters $a=-1$ and $b=c=0$.
$$\calL{GAN}(G_Y, D_Y) = 
\frac{1}{2} \E{y \sim p_{data}(y)}{(D_Y(x)-1)^2} + \frac{1}{2} \E{x \sim p_{data}(x)}{(D_Y(G_Y(x)))^2}$$
The second loss term corresponds to cycle-consistency losses for both directions.
$$\calL{cyc}(G_X, G_Y) = \E{x \sim p_{data}(x)}{||x-G_X(G_Y(x))||_1} + \E{y \sim p_{data}(y)}{||y-G_Y(G_X(y))||_1}$$
The loss of the final model sums all the partial loss terms with constant weights (regarded as hyperparameters).
$$\calL{}(G_X, D_X, G_Y, D_Y) = \calL{GAN}(G_X, D_X) + \calL{GAN}(G_Y, D_Y) + \lambda_{cyc}\calL{cyc}(G_X, G_Y)$$
The objective of training is summarized by the following optimization problem.
$$G_{X}^{*}, G_{Y}^{*} = \arg\min_{G_X, G_Y} \max_{D_X, D_Y} \calL{}(G_X, D_X, G_Y, D_Y).$$

\subsubsection{Conditioning on regions of interest}

To enhance the usefulness of our model, we add another input modality into to our generative model that represents regions of interest in the picture. For example, for breast cancer imaging, this modality could contain a boolean mask indicating segmented regions with ``suspicious'' (potentially cancerous) tissue. This also allows for encoding of various invariants into the dataset. By varying the additional mask position spatially, we obtain several variants of the transformed image, which together encode spatial equivariance of cancerous tissue, which might not be represented in the original dataset due to a low number of samples. The datasets we use all contain masks (of varying quality) with highlighted lesions or benign masses of the same dimension as the image.

To model the additional data source, we append another channel to our input image and let the model train using both the original image and the mask as both input and output. The generator now obtains a ``two channel'' image, and produces two channels instead of one. The final loss function is a sum of our $\cal{L}$ loss function applied to each channel individually. The rest of the model remains the same. The changes in the formulation of both generators and discriminators are the following (shown for $X \to Y$, $Z$ is the domain of masks):
\begin{align*}
G_{Y}&: (X, Z) \to (Y, Z)\\
D_{Y}&: (X \cup \hat{X}, Z) \to [0, 1]
\end{align*}

\subsubsection{Removing checkerboard artifacts}\label{sec:artifacts}

Empirically, models with deconvolutional layers tend to exhibit ``checkerboard'' artifacts, especially when trained for longer amounts of time \cite{odena2016deconvolution}. Therefore in our experiments, we 1) substitute a deconvolution with nearest-neighbor upsampling followed by a convolution, and 2) we initialize the kernel weights using ICNR \cite{icnr}. Generally, deconvolution preserves more details and produces less blurry results compared to upsampling and followed by convolution. We also evaluated bilinear upsampling, but it empirically produced more artifacts than nearest-neighbor upsampling.

\subsection{Neural classification model}

The classifier model used for all experiments was an adaptation of Faster R-CNN \cite{frcnn} that ranked second in the DREAM breast cancer detection challenge proposed by \citet{dezso}. Faster R-CNN is a convolution-based network capable of classifying and localizing objects in an image. Pure classification networks (predicting a binary answer) are easier to train, and thus more commonly used for mammography images. However, we believe that localizing malignant tumors is important if the system was to be implemented in clinical routine, since it helps in verifying the decision. The network is based on ResNet-50, a 50 layered network with residual connections pretrained on ImageNet \cite{imagenet}. Similarly to \citet{dezso}, we also changed the following parameters: we enabled the proposal network, and changed the proposal non-maximal suppression threshold to 0.5.

\section{Experiments}
To validate our ideas and claims, we propose several simple experiments in the domain of breast cancer recognition from 2D mammography images.

\subsection{Model implementation}

The generative augmentation models for all our experiments are based on the CycleGAN \cite{cyclegan} architecture, and are implemented in TensorFlow\footnote{Based on the TensorFlow research CycleGAN implementation: \url{https://github.com/tensorflow/models}} \cite{tensorflow}. More details about the architectures and training procedures are provided in Appendix~\ref{app:model}.

\subsection{Datasets}

There are several datasets that relate to breast cancer diagnosis. In most of these one can observe the limitations that we outlined in the introduction. For our experiments, we used the following datasets:
(1) BCDR \cite{bcdr}, Breast Cancer Digital Repository, several datasets from Portugal; (2) INbreast \cite{inbreast}, INbreast digital breast database, also from Portugal. Samples with a BiRads classification greater than 3 were considered as positive (cancerous), lower than 3 were considered negative (healthy); and
(3) CBIS--DDSM \cite{cbis}, Curated Breast Imaging Subset of DDSM (Digital Database for Screening Mammography) from the USA.

\begin{table}[btp]
\centering
\begin{tabular}{lrr}
\toprule
Dataset & Cancerous & Healthy\\
\midrule
BCDR-1 & 55 & 199\\
BCDR-2 & 44 & 651\\
INbreast & 100 & 270 \\
CBIS & 672 & 960\\
\bottomrule
\end{tabular}
\quad
\begin{tabular}{lrr}
\toprule
Dataset & Cancerous & Healthy\\
\midrule
Training & 655 & 1538 \\
Evaluation & 116 & 272 \\
Testing & 100 & 270\\
\bottomrule
\end{tabular}
\caption{Number of samples in the various datasets.}\label{tab:datasets}
\end{table}

For the generative model, we use BCDR-1 and BCDR-2 (merged together) for training. For the classifier, we use both BCDR datasets along with CBIS with an 85\% training and 15\% evaluation split. Due to a high noise ratio in CBIS, we only used it for the classifier. We use the held-out INbreast dataset as a test dataset for both models. All images were downscaled to $256 \times 204$ pixels due to hardware limitations. We also experimented with $512 \times 408$ pixels, but the image quality was poorer.
Table~\ref{tab:datasets} shows the number of samples in the respective datasets.

\subsection{Training a classifier}\label{sec:training}
Our Faster R-CNN \cite{dezso} based classifier was trained to localize malignant and benign lesions. We convert the pixel masks into a set of bounding box by applying Otsu threshold segmentation and taking the bounding box around every disconnected region. Images with no lesions or lesions with a bounding box area smaller than $10$ pixels were discarded, as R-CNN doesn't need to train on ``negative'' images. For each image, the model predicts a set of bounding boxes, corresponding scores, and classes. For evaluation, we treat an image as positive (cancerous) if any of the bounding boxes score with a malignant class is higher than a chosen, constant confidence threshold.

We train the classifier on different datasets for a maximum of 100,000 steps (batch size~8) and pick the best model based on ROC AUC \cite{roc} on the evaluation set. Based on inspection of the evaluation set loss, we empirically chose the models trained for $47,500$ steps (for all model variants).

\subsection{``Fooling'' a trained classifier}

As a first step, we want to see if our classifier, trained only on original images, is ``fooled'' by the generated images. In other words, for correctly classified images, how many times does the label change after we run the images through the generative augmentation model? We evaluate this question on all of our test data (see Section~\ref{sec:results}).
%See Figure~\ref{fig:fooling} for an illustration.

\begin{figure}[tb]
    \centering
    \includegraphics[width=0.85\textwidth]{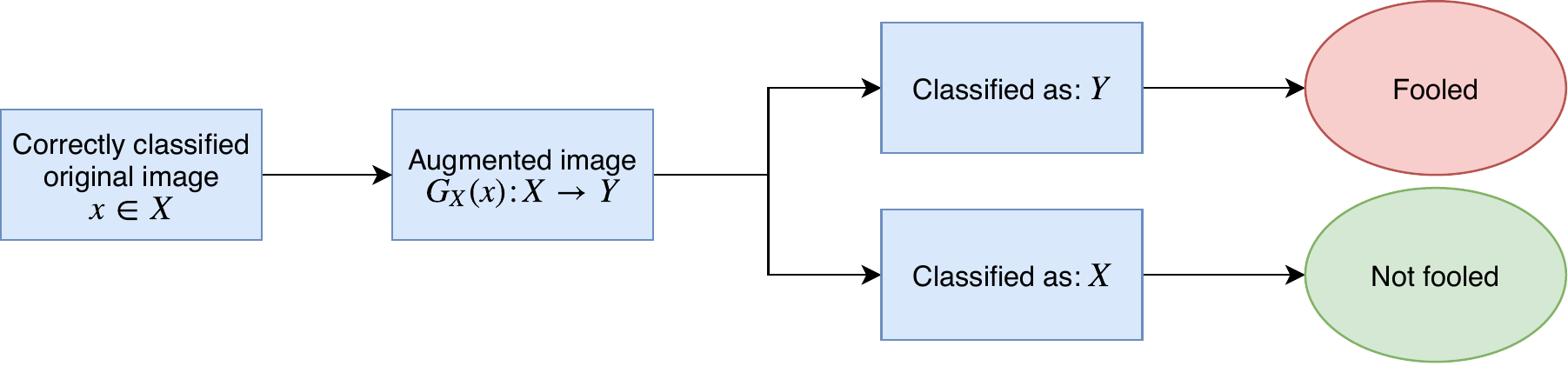}
    \caption{Evaluation diagram whether the generator $G_Y$ fooled the trained classifier into thinking that the generated image $G_Y(x)$ is from domain $Y$.}\label{fig:fooling}
\end{figure}

\subsection{Improving the classifier}

Secondly, we evaluate if a classifier trained in the same way on a mixed dataset of original and augmented images using the generative model performs better in terms of both classification metrics and ``being fooled'' We also compare the model to standard augmentation techniques such as image translation, rotation, and horizontal flipping. We use the same training/evaluation/testing split, but balance the training dataset by converting all the healthy images to cancerous ones, and adding them to our dataset. We then balance the dataset in a similar way as in Section~\ref{sec:training}.

\section{Results}\label{sec:results}

To visualize the results of our generative augmentation models, we show
a random uniform selection of images augmented by our generative model from the INbreast test dataset in Figure~\ref{fig:samples}~and~\ref{fig:samples_mask} (Appendix~\ref{app:figs}).

\begin{table}[tb]
\centering
\begin{tabular}{lrrrr}
\toprule
Classifier training data & Correctly clf.~\% & Fooled~\% & ROC AUC \% & F1 score \% \\
\midrule
Original & $76.22 \pm 4.08$ & $38.49 \pm 3.31$ & $83.50 \pm 1.47$ & $62.53 \pm 0.40$ \\
Classically augmented & $80.54 \pm 0.47$ & $33.34 \pm 2.53$ & $79.05 \pm 1.94$ & $62.63 \pm 2.50$ \\
GAN-augmented & $80.99 \pm 1.96$  & $30.91 \pm 8.83$ & $82.04 \pm 0.57$ & $63.81 \pm 2.04$ \\
\bottomrule
\end{tabular}
\caption{Fooling and improving the classifier evaluated on the test dataset INbreast (different patient population than the training set). GAN-augmented images are from the unconditioned GAN model because of better image quality. Each run was repeated three times --- shown are the average and the standard deviation for each value.}
\label{tab:results}
\end{table}

The first and second columns of Table~\ref{tab:results} show that the classifier learns to be less fooled by our generative augmentation model if we augment the training set images using the same model, which confirms the intuition that this makes the classifier slightly more robust.

As shown in the first row of Table~\ref{tab:results}, the classifier performs reasonably well when trained on the original dataset and evaluated on a test split from that dataset (both in terms of ROC AUC and F1 score). The F1 score is computed using a custom bounding box proposal confidence threshold of $0.23$, same as in \citet{dezso}.

When the training set images are augmented by our GAN (third row), the average ROC AUC goes down slightly, but the error margin is too big to produce a conclusive result. 
As was previously shown by \citet{rsna_abstract, bgan2} and our subjective assessment, this suggests that the new GAN-generated data might be challenging to classify for our classifier.
The same conclusion applies for the experiment where we augment the training set images using traditional image augmentation techniques.

\section{Discussion}

Overall, our GAN training has been very prone to checkerboard and ``S''-shaped artifacts, as can be partially seen in Figures~\ref{fig:samples} and \ref{fig:samples_mask} (Appendix~\ref{app:figs}). We also experimented with both higher~($512 \times 408$~px) and lower resolutions~($256 \times 208$~px) of images: the lower resolutions generally had fewer artifacts and faster training times, but a higher resolution is desirable when thinking about moving to full-field mammographic images in the future. Unfortunately, due to GPU memory limitations resolution could not be further increased. Our GAN models and RCNN-based classifiers train in less than 24 hours on an NVIDIA TITAN Xp GPU.

The classifier results are inconclusive, and it is not clear that adding our augmented images helps the classifier achieve better performance or not. We hypothesize that this might be due to the noise in our data, as the results of \citet{lesionremoval} suggest that the overall method is sound and can improve classifier performance if applied well.

\section{Future work}

Possible future improvements to our work include investigating upscaling the resolution without obtaining artifacts with approaches similar to \citet{p2pHD}, stabilizing the conditioned model training and results, and also leveraging that model fully to augment the images in pre-specified places.  For a more detailed image, we could explore approaches similar to Self-Attention GAN \cite{selfattention}, which promises to pay close attention to parts of the input image for output generation. This would also help in interpreting the resulting changes done by the GAN. Unfortunately, this approach is very memory-expensive.

Traditionally, Variational Autoencoders \cite{vae} (VAEs) lack detail in the output images and GANs lack ``truthfulness'' --- they may overgenerate parts of the image \cite{gansequal}. As a more hybrid approach, we could combine a VAE with a GAN to model both the location and the image details jointly with one model, similarly to the approaches in \cite{unit, munit, basel}. To simplify the model, one could also try using a StarGAN-like \cite{stargan} approach by only using one generator/discriminator pair which is conditioned by the class label, instead of using two generators and discriminators.

\section{Conclusion}

In our work, we have shown that for binary classification on images, there exists a simple way to potentially increase prediction accuracy by generative dataset augmentation. Leveraging the idea behind CycleGAN, we have designed a GAN that is able to translate images from one class label to the other, and use that property to augment the training dataset of a classifier into a bigger, more balanced, and less sparse dataset. We have provided a proof of concept implementation and shown that on the challenging noisy example case of breast cancer recognition from mammography images, we may be able to help improve performance of classifiers. This suggests our generative augmentation model learns a meaningful approximation of the manifolds of our class labels.

\midlacknowledgments{We would like to thank the Computer Vision Lab at ETH Z\"{u}rich for providing us with computational resources.}

\bibliography{references}

\newpage

\appendix

\section{Model implementation}\label{app:model}

We train all our GAN models for $40,000$ steps, using a learning rate of $1 \cdot 10^{-4}$ for the discriminators and $2 \cdot 10^{-4}$ for the generators. The optimization is performed using Adam \cite{adam} and a batch size of 1. All code is available on GitHub\footnote{\url{https://github.com/BreastGAN/augmentation}}.

The architectures of both discriminators are the same: 4 convolutional layers with reflection padding, with filters of size 64, 128, 256, 512 and stride 2 for all layers except for the last one that has stride 1, with a LeakyReLU activation function \cite{relu, relu2, leakyrelu}: $\max(0.2x, x)$. All the convolutions have a kernel size of $4 \times 4$. The output is subsequently flattened to one channel using a stride 1 convolution, with a sigmoid activation function.

Both generator networks consist of two convolutions with stride 2 to compress the dimensionality of the image followed by 9 ResNet blocks (2 convolutions layers each). Lastly, the result is upsampled using two additional convolutional layers as described in Section~\ref{sec:artifacts}. All the generator layers use ReLU activation functions.

\newpage

\section{Random samples from our GAN augmentation models}\label{app:figs}

\begin{figure}[htbp]
    \centering
    \subfigure[Healthy (top) to cancerous (bottom).]{
        \centering
        \includegraphics[width=0.9\textwidth]{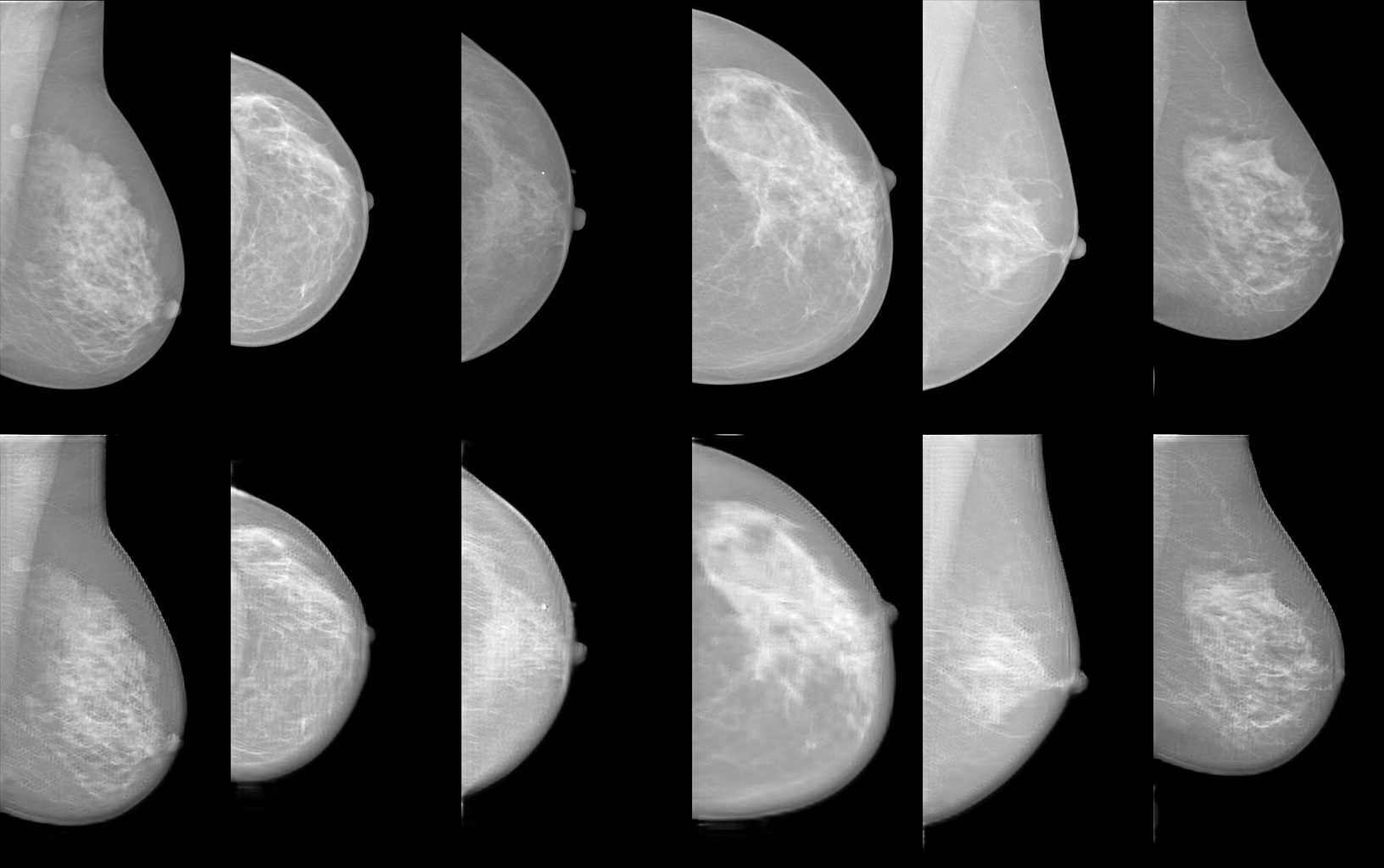}
    }
    
    \subfigure[Cancerous (top) to healthy (bottom).]{
        \centering
        \includegraphics[width=0.9\textwidth]{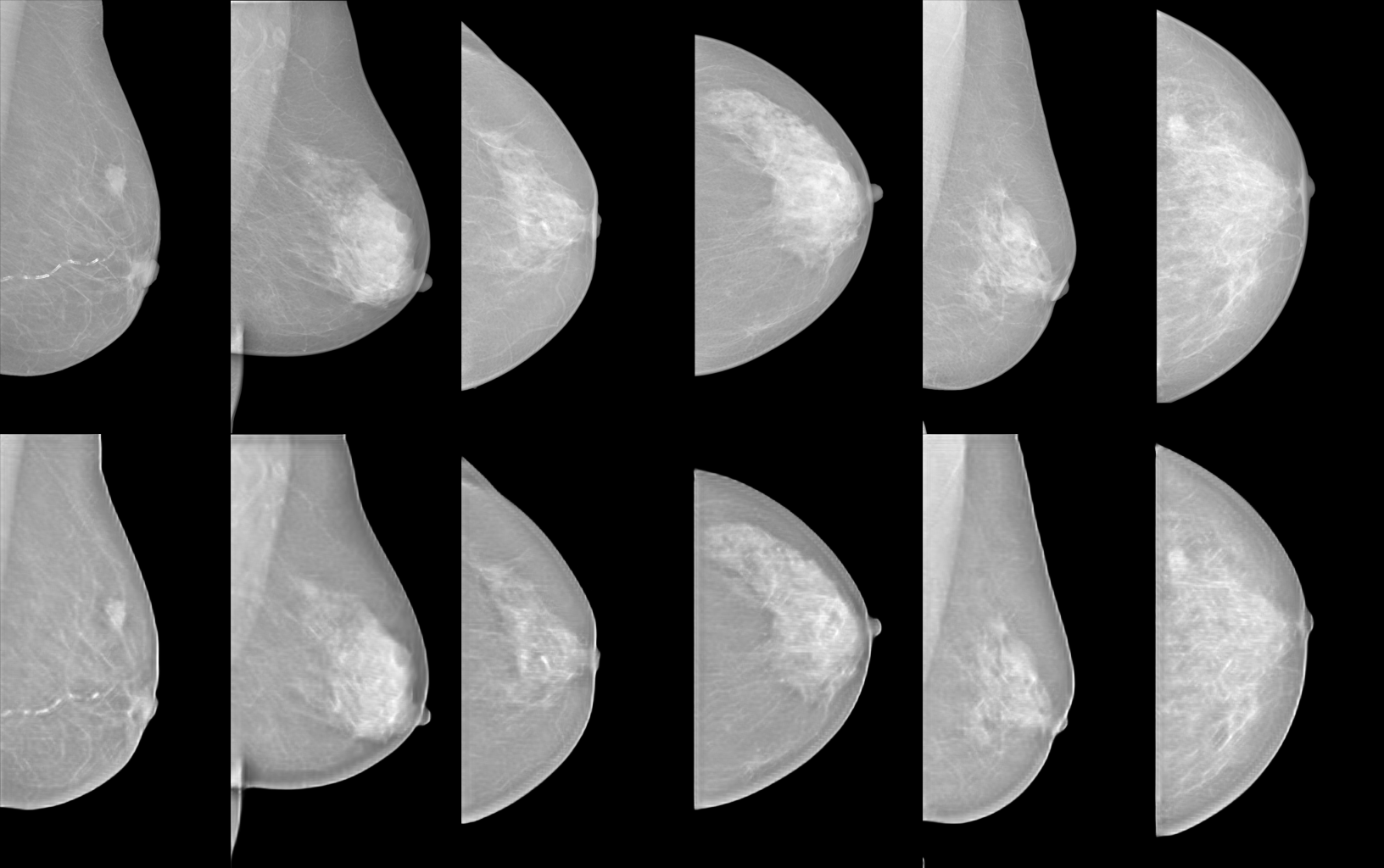}
    }
    \caption{Random samples of images from our trained GAN (without masks, $512 
\times 408$ px).}\label{fig:samples}
\end{figure}

\begin{figure}[htbp]
    \centering
    \subfigure[Healthy (top) to cancerous (bottom), mask (middle).]{
        \centering
        \includegraphics[height=0.42\textheight]{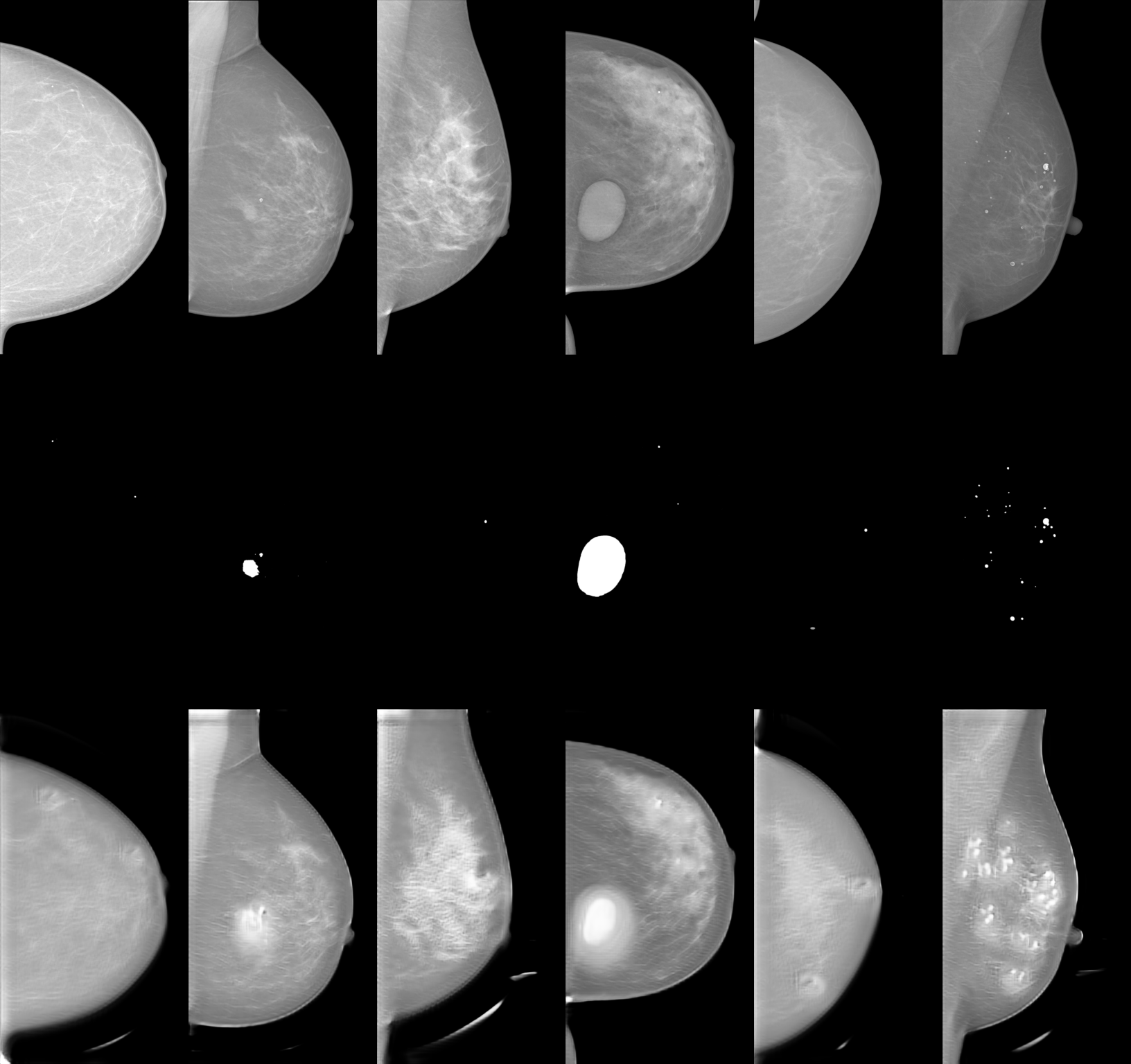}
    }
    
    \subfigure[Cancerous (top) to healthy (bottom), mask (middle).]{
        \centering
        \includegraphics[height=0.42\textheight]{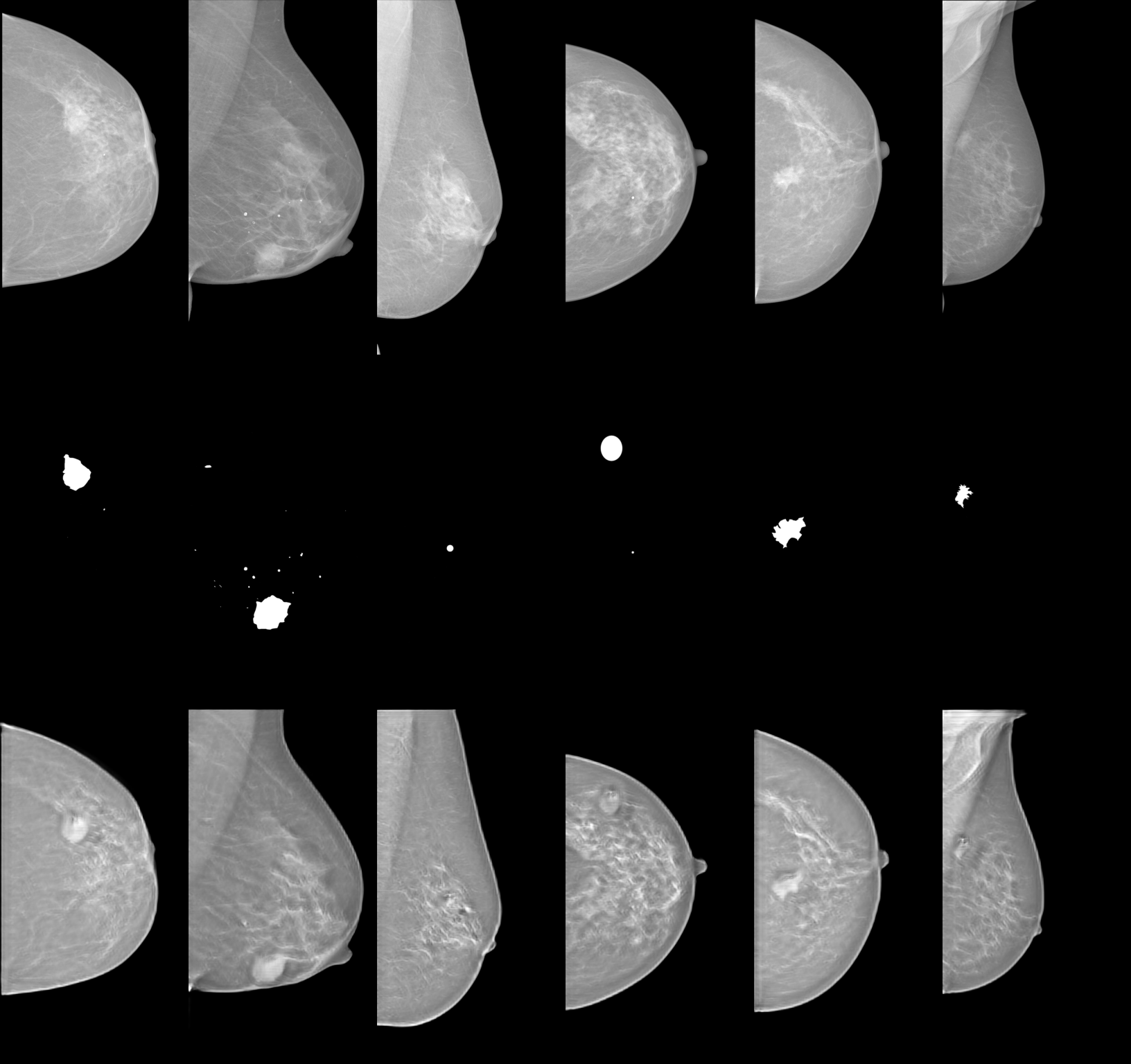}
    }
    \caption{Random samples of images from our trained GAN (with masks, $512 
\times 408$ px).}\label{fig:samples_mask}
\end{figure}

\end{document}